# Dental Severity Assessment through Few-shot Learning and SBERT Fine-tuning

Mohammad Dehghani*
Electrical and Computer Engineering
Department,
University of Tehran,
Tehran, Iran
dehghani.mohammad@ut.ac.ir

**Abstract**

Dental diseases have a significant impact on a considerable portion of the population, leading to various health issues that can detrimentally affect individuals' overall well-being. The integration of automated systems in oral healthcare has become increasingly crucial. Machine learning approaches offer a viable solution to address challenges such as diagnostic difficulties, inefficiencies, and errors in oral disease diagnosis. These methods prove particularly useful when physicians struggle to predict or diagnose diseases at their early stages. In this study, thirteen different machine learning, deep learning, and large language models were employed to determine the severity level of oral health issues based on radiologists' reports. The results revealed that the Few-shot learning with SBERT and Multi-Layer Perceptron model outperformed all other models across various experiments, achieving an impressive accuracy of 94.1% as the best result. Consequently, this model exhibits promise as a reliable tool for evaluating the severity of oral diseases, enabling patients to receive more effective treatment and aiding healthcare professionals in making informed decisions regarding resource allocation and the management of high-risk patients.

## I. INTRODUCTION

The incidence of periodontitis and dental caries has witnessed a surge in recent years among the human population, highlighting the pressing need for early detection to prevent severe complications and tooth loss [1]. Dental caries is a significant health concern affecting both children and adults in most industrialized nations [2]. Its impact is felt throughout an individual's lifetime, leading to pain, discomfort, and oral deformities. Furthermore, research has established a clear association between dental issues and other major diseases, including cancer, cardiovascular disease, diabetes, and chronic respiratory disease [3].

The advent of cone-beam computed tomography (CBCT) systems has brought about a transformative impact on the field of oral and maxillofacial radiology [4]. CBCT serves as an imaging technique that enables the acquisition of three-dimensional representations of teeth or dentition in relation to the surrounding skeletal structures [5]. The utilization of CBCT in dental settings offers numerous advantages, including its compact size, affordability, and low exposure to ionizing radiation, which have contributed to its widespread adoption in dental practice as compared to medical computed tomography [4]. The applications of CBCT imaging encompass various areas such as restorative dentistry, surgical procedures, tooth localization, implant treatment planning, and assessment of bony pathologies in the jaws [6]. Hence, CBCT imaging allows for a comprehensive assessment of disorders affecting the oral cavity, jaw, and facial region, necessitating the development of automated tools for their analysis.

The utilization of artificial intelligence (AI) has experienced a significant surge across various industries in recent years, including the field of dentistry, where it has proven instrumental in making complex predictions and facilitating decision-making processes [7]. The application of AI in dentistry enables more efficient, accurate, and time-saving diagnosis and treatment planning. One notable task that can be achieved through AI is the classification of intraoral periapical radiographs, where deep learning networks can be employed to accurately identify the location of teeth [8]. Deep learning techniques, mostly integrated with image processing methods, have been successfully utilized in classifying different pathologies and anatomical structures, such as maxillary sinusitis [9], changes in the temple-mandibular joint [10], and root morphology in molar teeth [11].

Text classification has recently gained significant attention in the medical field [12]. With the advancements in big data and analytics, healthcare professionals are leveraging this opportunity to diagnose and analyze available data, leading to improvements in medical services. Natural language processing (NLP) techniques enable the analysis of extensive patient datasets, which are then structured into training datasets to educate automated systems. A variety of applications of NLP in medicine are being explored, including the classification of medical prescription [13], medical question answering [14, 15], and analyzing biomedical literature [16].

In the dental industry, the widespread use of Electronic Dental Record (EDR) systems has made them an essential component of research aimed at using data to inform clinical decisions [17]. EDR systems contain a wealth of valuable information about dental patients, largely in the form of unstructured text [18]. However, there have been few attempts to apply NLP to dental issues. Using NLP, radiology reports can be automatically analyzed to provide valuable information about patients' health and diseases [19].

The aim of this study is to evaluate and compare different machine learning/deep learning methods and large language models (LLMs) to assess the severity of dental conditions. The objective is developing an automatic diagnosis technology that can assist even experienced dentists by reducing their workload and increasing efficiency. Several algorithms were compared, including Multinomial Naive Bayes, Gaussian Naive Bayes, Decision Tree, Random Forest, Gradient Boosting, Linear Support Vector Machine (SVM), Radial Basis Function (RBF) kernel SVM, Logistic Regression, Multi-Layer Perceptron (MLP), combination of Convolutional Neural Networks (CNN) and Bidirectional Long Short-Term Memory (BiLSTM), Few-shot learning with SBERT and MLP (FSBM), and GPT-3.5.

Based on the provided radiology dataset, the following scenarios were considered:

(1) **Urgent attention required:** This scenario includes cases where immediate medical attention is necessary, such as mouth and jaw tumors,
(2) **Treatment required but can be delayed:** In this scenario, there are serious dental problems that require treatment, but the medical attention can be delayed without significant risk,
(3) **Optional treatment:** This scenario involves cases where there is an existing dental issue, but the need for treatment is subjective and depends on the individual's preference,
(4) **No problem detected:** This scenario indicates that no dental issues were identified in the radiology dataset.

By assessing the severity of dental conditions and categorizing them into these scenarios, the automatic diagnosis technology can provide valuable guidance to dentists, helping them prioritize cases and optimize their workflow.

The remainder of the paper is organized as follows. A review of related works is presented in Section 2. A description of the applied model is provided in Section 3. The results of the study are presented in Section 4 along with a comparison of the performance of various models. In section 5, the paper concludes.

## II. RELATED WORK

There has been an increasing use of AI in dentistry, including operative dentistry, periodontics, orthodontics, oral and maxillofacial surgery, and prosthodontics. AI applications in dentistry primarily focus on radiographic or optical imaging diagnosis [20]. In the field of dental diagnostic images, a number of popular deep learning tasks have been applied.

Ahmed et al [21] assessed the ability of U-Net architecture to detect and classify dental caries using ResNet-50, ResNext101, and Vgg19 networks. Data was collected from the Electronic Medical Record System at KAU Dental Hospital in Saudi Arabia. An object's location, size, and shape in each image were determined using semantic segmentation. A total of 554 bitewings were collected and labeled using RoboFlow. In the overall model, the IoU score is 55.1%, while the F1-score is 53.5%.

Setzer et al. [22] presented an automated approach to segmenting CBCT images and detecting periapical lesions using a U-Net architecture. The segmentation process resulted in voxels being labelled as "lesion" (periapical lesion), "tooth structure," "bone," "restorative material," and "background." According to the evaluation, the accuracy of lesion detection was 93% and specificity was 88%.

Kim et al. [23] have developed an algorithm that detects landmarks in lateral cephalogram images using stacked hourglass deep learning models. A dataset containing 2,075 lateral cephalograms and ground truth positions of 23 landmarks was constructed, and the algorithm provided a classification rate of 88.43%.

In the study conducted by Bianchi et al. [24], an analysis of 52 clinical, biological, and high-resolution CBCT markers was presented with the aim of early diagnosis of Temporomandibular Joint (TMJ) disorder. The researchers employed logistic regression, random forest, LightGBM, and XGBoost algorithms to train the prediction model. It has been determined that the XGBoost + LightGBM model was capable of achieving accuracy of 82.3%, AUC of 87%, and F1-score of 82.3% in order to diagnose the disease.

Although a wide range of computer vision-based methods have been studied in this area, but research on NLP has been limited. The potential of using NLP in conjunction with EDRs to automate the classification, extraction, and retrieval of information from clinical notes pertaining to dental patients has yet to be fully explored [18].

According to Ohtsuka et al. [25], NLP can be used to automate orthodontic diagnosis. This study aimed to enhance automated diagnosis using concise summaries of medical findings written by dentists. Various techniques, including Bag of Words and Support Vector Machines (BOW+SVM), Recurrent Neural Networks (RNN), CNN, Self-Attention Networks, and Bidirectional Encoder Representations from Transformers (BERT), were employed to train the model. An evaluation of 970 Japanese medical findings indicated that CNN exhibited the highest performance, achieving an F1-score of 44% for the original documents and 48% for the short summaries.

Shimizu et al. [26] conducted a study concentrating on orthodontic diagnosis and treatment planning. The study utilized documents containing clinical information, a prioritized list, and treatment strategies for 967 cases. To process the text data, the BOW method was employed to convert the documents into vectors. The researchers developed their proposed models using a combination of a SVM and a self-attention network. The clinical evaluations involved the participation of four orthodontists who assessed the developed models and their performance.

Chen et al. [27] presented a NLP workflow aimed at extracting information from EDRs for utilization in clinical decision support systems. The workflow involved utilizing attributes, attribute values, and tooth positions present in the EDRs. To facilitate this, the researchers employed Sentence2vec and Word2vec algorithms for learning sentence vectors and word vectors, respectively.

In their study, Patel et al. [28] utilized longitudinal EDRs to extract information from clinical notes with the objective of classifying patients into three groups based on periodontal disease change: disease improvement, disease progression, and no change. The study leveraged a dataset of 28,908 patient records obtained from the Indiana University School of Dentistry (IUSD) clinics.

Huang et al. [29] explored the potential applications of LLMs in the field of dentistry. The authors focus on two main LLM methods: automated dental diagnosis and cross-modal dental diagnosis, and discuss the potential uses and implications of each.

### III. METHODOLOGY

The applied method follows a general overview, as depicted in Fig. 1. The process begins with data collection, where relevant information is gathered for analysis. This data is then preprocessed to prepare it for further analysis and modeling.

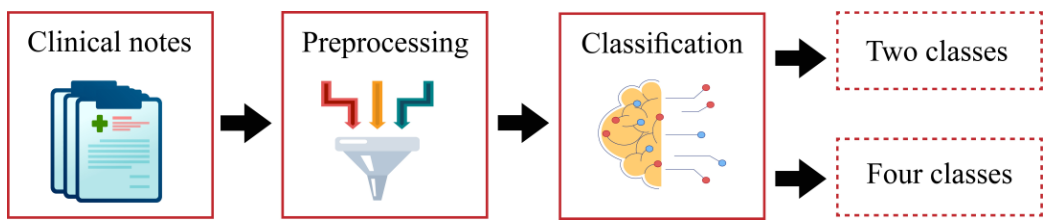

Fig. 1: Overview of the applied method.

#### 1. Dataset

In order to conduct this study, a total of 704 records from the dental radiology database [30] were analyzed. The reports were categorized into four distinct groups based on their content and characteristics, as outlined in Table 1. To provide a better understanding, Table 2 presents a sample dataset for each class, illustrating examples from each category. This sample dataset demonstrates the diversity and representation of the different classes within the analyzed records.

Table 1: The explanation of each class.

| Class | Explanation |
|---|---|
| 1 | Issues require urgent attention |
| 2 | Treatment can be delayed |
| 3 | The problem is not urgent (optional treatment) |
| 4 | Conditions are entirely normal (no treatment require) |

Table 2: Example of each class.

| CBCT of the patients | Label |
|---|---|
| There is a mixed lesion with poorly-defined border in anterior part of the mandible. It caused loss of continuity of the buccal and lingual cortices and alveolar crest.<br>Based on the possible differential diagnoses (DDX):Osteomyelitis in the healing site of the previous surgery, Infected fibro-osseous lesion, R/O Sarcomatosis lesions such as chondrosarcoma | 1 |
| The tooth # 8 is in inverted fashion. The tooth is tightly attached to the buccal and palatal cortices. Loss of continuity of the buccal and palatal cortices is seen. The blunting in 1/3 middle of the root of tooth # 7 is detected. There is association (contact without any cortex) between incisive canal and tooth # 8. | 2 |
| There is a mesiodenses (MD) invertedly positioned in the palatal side of the tooth #8. The MD caused loss of continuity of the palatal cortex. No root resorption of the adjacent teeth is found. There is no other supernumerary or missing tooth in the jaws. | 3 |
| No erosive lesion can be detected in ant. maxilla. | 4 |

## 2. Data preprocessing

In order to perform text classification tasks, data preprocessing is essential. In this process, raw data is transformed into a clean, organized, and suitable format for modeling. We implemented several necessary preprocessing on the text descriptions, including:

- **Tokenization:** Text is divided into words or terms [31].
- **Removing stop words:** Removing high-frequency words that possess limited semantic value or contribute minimal information to the text classification task [32].
- **Lemmatization:** A word from its inflectional form to its base or lemma form [33].
- **Spell error correction:** Within the domain of medical manuscripts, utmost importance is placed on precision and accuracy. Considering the origin of our dataset from such documents, a meticulous focus was dedicated to the pre-processing stage to rectify any discrepancies and uphold the integrity and quality of our analysis.
- **Random oversampling:** It is the simplest technique for oversampling that minority target instances are duplicated randomly to achieve a more balanced distribution of classes [34, 35].

## 3. The models used in this study

During the modeling phase, a pivotal procedure entailed the transformation of input textual data into numerical representations, known as feature vectors. To accomplish this, our approach leverages the Term Frequency-Inverse Document Frequency (TF-IDF) technique for feature extraction from sentences. TF-IDF is a methodology employed to ascertain the significance of a term within a document, determined by the term's frequency and inversely proportional to the frequency of its occurrence across the entire corpus, thereby establishing the term's relevance. [36, 37]. The input data is fed into several classification algorithms, which serve as baseline models. The subsequent description outlines these algorithms employed for training distinct models.

*Multinomial Naive Bayes*: In order to classify discrete features (for example, word frequency for text classification), the Multinomial Naive Bayes algorithm is applied. While integer feature counts are required for multinomial distributions, fractional values may also be appropriate in cases such as TF-IDF [38, 39].

*Gaussian Naive Bayes:* Gaussian Naive Bayes is a derivative of the Naive Bayes algorithm that incorporates the Gaussian normal distribution for classification purposes and is specifically designed to handle continuous data. This algorithm assumes that the features employed for classification follow a Gaussian likelihood distribution [40].

*Decision Tree:* The decision tree, based on the principle of divide-and-conquer, is a recursive learning method. Beginning at the root node, the dataset is partitioned into various subsets based on specific criteria. The decision tree structure comprises leaf nodes that represent distinct categories and child nodes that represent subsets of the dataset [41].

*Random Forest:* A random forest is constructed by the aggregation of multiple trees. As the number of trees in the forest increases, the generalization error of the ensemble tends to decrease owing to the individual strength of each tree and their interrelationships. This collective behavior of the random forest, facilitated by the combination of diverse trees, contributes to enhanced predictive performance and improved generalization capabilities [42].

*Gradient Boosting:* The gradient boosting algorithm is based on the concept of ensemble learning. By combining several simple learning algorithms, the algorithm achieves better predictions than any of the individual algorithms [43].

*SVM:* SVM aims to construct hyperplanes within one-dimensional input spaces or feature spaces by maximizing the margin, which represents the distance between the hyperplanes. This optimization process facilitates the attainment of optimal generalization capabilities. Ideally, the category boundary should be positioned as far as possible from the hyperplane, along the direction perpendicular to it, ensuring a clear separation between different classes [37].

*Logistic Regression:* In logistic regression, individual predictors are assigned coefficients that quantify their independent contributions to the variability of the dependent variable [44]. Logistic regression assumes a non-linear relationship between the independent variables and the dependent variable, with the logit of the outcome being associated with the predictor values [45].

*MLP:* MLP often referred to as a "vanilla" neural network, is a simple neural network architecture frequently employed for automated feature extraction. The model comprises three layers: an input layer, a hidden layer, and an output layer. Each node within the network is assigned a specific weight, which determines its influence on the overall computation and information flow [46].

*CNN+BiLSTM:* As depicted in Fig. 2, the initial deep learning model applied in this study incorporates a combination of CNN and BiLSTM. This hybrid architecture enables the processing of textual data sequences by leveraging the capabilities of both convolutional and sequential models. The convolutional extracts features from the input data, which are subsequently fed into the sequence model. In order to extract new chronological features of a text, the sequence model treats the produced features from the convolution as sequences [47].

Within the CNN layer, distinctive patterns that characterize the textual data are extracted, thereby capturing important text features. Subsequently, a pooling technique based on maximum values is employed, enabling the mitigation of any adverse impact caused by the presence of PAD tokens (commonly known as padding) appended to shorter sentences. This pooling approach ensures that the influence of the padding tokens is minimized, thereby preserving the integrity of the extracted features [48]. Subsequently, the model undergoes training using a BiLSTM layer. By leveraging the Long Short-Term Memory (LSTM) architecture, an improved version of RNN, the issue of vanishing gradients is effectively mitigated. The BiLSTM design encompasses a specialized cell capable of storing information across arbitrary time intervals, along with three gates that regulate the flow of information within the network [49, 50]. The utilization of BiLSTM enables the retention of semantic information from both preceding and subsequent tokens in the sequence. To enhance regularization and prevent overfitting, a dropout layer is employed, selectively discarding 10 percent of the data. Next, a dense layer is applied to generate the final output.

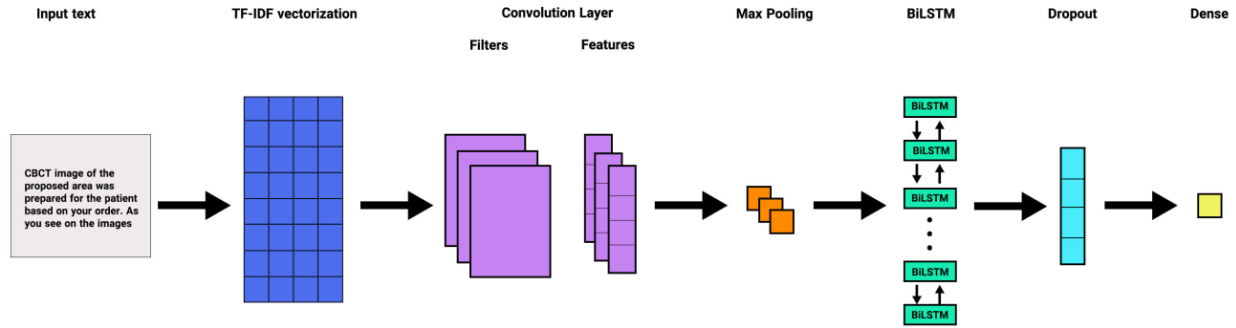

Fig. 2: Applied CNN+BiLSTM model.

*GPT-3.5:* LLMs pertain to a specific category of pretrained language models obtained by augmenting the model size, pretraining corpus, and computational resources [51]. The research community commonly labels GPT-3 and its successor large-scale models as "Large Language Models" to distinguish them from smaller-scale Pretrained Language Models (PLMs). This nomenclature is employed to emphasize the substantial difference in model size, capacity, and computational resources between these large language models and their smaller counterparts [52]. The GPT-3 and GPT-3.5 language models developed by OpenAI are designed to generate natural language text that is similar to the text generated by human beings [53].

*In-context learning (ICL):* ICL is a recent technique in NLP for LLMs, like GPT-3 [54], that allows models to adapt and improve their performance over time as they encounter new information. Moreover, similar to human decision-making processes, it undertakes new tasks using prompts to predict unseen samples.

*Zero shot learning (ZSL):* It was introduced as an approach to classify categories that are absent from the training set [55]. This method exhibits an inherent capacity to automatically recognize unseen objects, closely resembling the cognitive system of humans. ZSL is specifically formulated to facilitate the transfer of learned tasks from seen classes to unseen classes, enabling the natural recognition of novel objects belonging to previously unseen categories [56].

*Few-shot learning (FSL):* Similar to ZSL, but with a distinction in the availability of a limited number of labeled examples per class, the objective of this approach is to make predictions for new classes based on a sparse set of labeled data. The goal is to effectively identify and classify new objects with only a few samples available for training [57].

In order to augment the diversity and intricacy of prompts employed, we devised a methodology for generating a range of prompts with varying levels of difficulty. This was accomplished by harnessing the capabilities of the GPT model in conjunction with FSL techniques. By adopting this approach, we were able to evaluate the performance of the model across a spectrum of linguistic complexities. Such comprehensive testing provided valuable insights into the model's adaptability and resilience in effectively processing prompts of different types. Table 3 shows example of GPT-3.5 prompts.

Table 3: Prompts for second stage task.

| | |
|---|---|
| **Simple prompt** | Assess the magnitude of dental and oral health concerns by analyzing this clinical note. Provide a numerical label (0 or 1), where:<br><br>0: urgent<br>1: non-urgent<br><br>note: <example 1><br>label: 0<br>note: <example 2><br>label: 1<br>note: <text><br>label: |
| **Complicated prompt** | We have images related to Cone-beam computed tomography systems (CBCT) and a report is prepared for each image. Now we want to determine how bad the patient's condition is based on these reports.<br>For example, when a patient's report is as follows "CBCT image was prepared for the patient based on your order. As you see based on the images: There is a mixed lesion with a poorly-defined border in the anterior part of the mandible. It caused loss of continuity of the buccal and lingual cortices and alveolar crest. DDX: Osteomyelitis in the healing site of the previous surgery Infected fibro-osseous lesion R/O Sarcomatosis lesions such as chondrosarcoma"<br>Due to the presence of mixed lesion, the patient's condition is not good and he should be treated immediately In other words, the condition of the patient is emergency.<br>But when a patient's report is as follows: "Based on CBCT images: As you see in images no erosive lesion can be detected in ant. maxilla."<br>This means that the patient is in a good condition and does not need urgent attention because no erosive lesion has been observed in the patient's report.<br><br>According to the above examples, receive the patient's text report as input and provide a numerical label (0 or 1), where:<br><br>0: urgent<br>This designation is assigned when there is a critical and time-sensitive issue that requires immediate attention and intervention. Examples of situations warranting a label of "0" include significant complications, potential risks to the patient's health, or conditions that may rapidly worsen if not addressed promptly.<br><br>1: non-urgent<br>This label is assigned when the observed issues, while noteworthy, do not pose an immediate threat to the patient's health and can be addressed over time with monitoring or future intervention. It indicates that the situation does not demand urgent action but may still require attention, treatment, or follow-up care in the long run. |

| | note: <example 1> |
| | label: 0 |
| | note: <example 2> |
| | label: 1 |
| | note: <text> |
| | label: |

*FSBM:* To enhance the efficacy of FSL using Sentence Transformers, we employed the SetFit framework [58], specifically tailored for fine-tuning Sentence Transformers in a FSL scenario. Unlike conventional methods reliant on prompts, SetFit adopts a distinct approach by directly generating comprehensive embeddings from textual examples. This methodology facilitates more efficient and effective training. Importantly, SetFit operates without the requirement of large-scale models such as GPT-3, offering a streamlined and resource-efficient alternative.

SetFit has gained recognition for its capability to achieve remarkable accuracy even when confronted with a scarcity of labeled data, rendering it a well-suited solution for scenarios characterized by sparse annotation. In our fine-tuning procedure employing SetFit, we extracted 200 samples from each class within the training dataset. These samples were utilized as the basis for generating pairs of sentences, with the objective of constructing a robust dataset that would facilitate subsequent modeling and fine-tuning processes.

To leverage the potency of semantic representations, we employed an SBERT-based network to process the generated pairs of sentences. The SBERT architecture, introduced by Reimers and Gurevych [59], builds upon the pretrained BERT network while incorporating siamese and triplet network structures. During the fine-tuning phase, the model acquired the ability to extract significant semantic information from the paired sentences, enabling a more nuanced understanding of their underlying meaning.

The fine-tuned SBERT model generated sentence embeddings for each input sentence, encapsulating semantic information within their vector representations. These embeddings facilitated effective comparison using cosine similarity, enabling a robust assessment of the semantic similarity between sentences. Ultimately, these embeddings were utilized as input to a classifier, such as a MLP, enabling informed decision-making based on the underlying semantic structure of the sentences. Fig. 3 illustrates SetFit framework. As a matter of fact, we applied the FSBM method to dentistry for the first time.

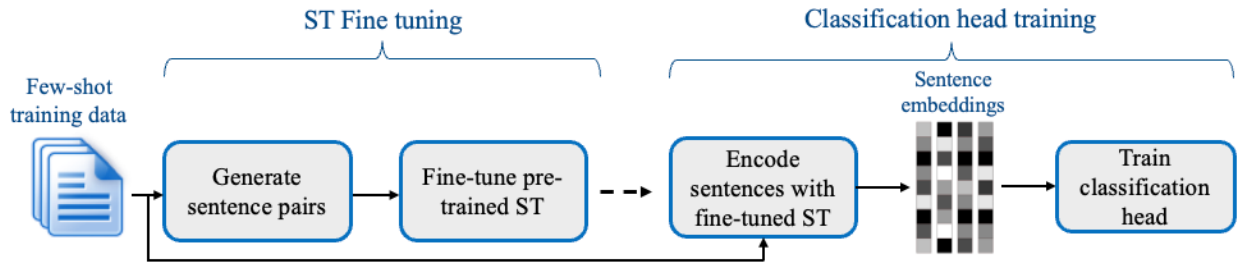

Fig. 3: SetFit Framework [58].

## IV. RESULTS AND EVALUATIONS

The impact of imbalanced data, characterized by skewed class distributions, on the accuracy of various classification algorithms is widely recognized [59]. In the medical domain, data imbalance is a prevalent issue, and the application of class imbalance methods [60] has been extensively employed. Typically, datasets exhibit a higher proportion of normal samples compared to abnormal samples, resulting in a substantial disparity between the two [61]. To mitigate potential bias and ensure a balanced representation of data across all classes, we employed a Random oversampling technique [63]. This approach aimed to address the variance in sample sizes among the four classes, as indicated in Tables 4 and 5. By oversampling, all members of the minority and majority classes were included, ensuring that no information from the original training set was lost. This oversampling technique offered the advantage of maintaining the integrity of the original dataset while mitigating class imbalance [62].

In this study, two stages of experimentation were conducted:
(1) **First stage:** The dataset comprises four distinct classes, as detailed in Table 1. The distribution of samples at this stage is illustrated in Table 4. Initially, the number of original samples for the first, second, third, and fourth classes were 67, 354, 219, and 64, respectively, yielding a total of 704 samples. Notably, the first and fourth classes exhibited fewer samples compared to the other two classes. To address this class imbalance, the data was balanced by obtaining 354 samples for each class, resulting in a total of 1416 samples after the balancing procedure.
(2) **Second stage:** At this stage, the dataset is divided into two distinct classes. The first class combines samples labeled as 1 and 2, representing patients requiring compulsory treatment. The second class comprises samples labeled as 3 and 4, representing patients who do not require compulsory treatment. The distribution of samples for this stage is presented in Table 5. Specifically, the first class contains 421 samples, while the second class contains 283 samples. To address any class imbalance, the number of samples in each class is balanced, resulting in a total of 708 samples for both classes after the data balancing process.

Table 4: Distribution of classes in first stage.

| Class | Number of samples (imbalance) | Number of samples (balance) |
|---|---|---|
| 1 | 67 | 354 |
| 2 | 354 | 354 |
| 3 | 219 | 354 |
| 4 | 64 | 354 |
| Total | 704 | 1416 |

Table 5: Distribution of classes in second stage.

| Class | Number of samples (imbalance) | Number of samples (balance) |
|---|---|---|
| 1 | 421 | 708 |
| 2 | 283 | 708 |

The dataset is divided 80/20, with 80% of the data being used to train models and the remaining 20% being used for testing. In order to compare results, we considered the results of a dataset that is divided into two or four categories, as well as whether the dataset is presented in a balanced or imbalanced manner.

While analyzing the dataset mentioned in reference [30], we came across a significant challenge stemming from a fundamental theoretical misinterpretation discussed in this reference. The authors treated the text classification task as sentiment analysis and carried out their work under this incorrect assumption. As a result, it was not possible to directly compare their findings with our work.

The results of the applied model for the first stage, where the data are imbalanced, are shown in Table 6. Based on the results, FSBM model outperformed other models with accuracy of 71.7%, precision of 70.2%, recall of 71.6%, and F-measure of 70.8%. Table 7 presents results in first stage, but the data is balanced. An accuracy of 89.1%, precision of 88.1%, recall of 88.5%, and F-measure of 88.2% were achieved by the FSBM model.

Table 6: First stage results for imbalanced data.

| Classifiers/ Evaluation metrics | Accuracy | Precision | Recall | F-measure |
|---|---|---|---|---|
| Gaussian Naive Bayes | 0.585 | 0.598 | 0.585 | 0.583 |
| Decision Tree | 0.500 | 0.518 | 0.500 | 0.507 |
| Gradient Boosting | 0.656 | 0.656 | 0.660 | 0.641 |
| Random Forest | 0.642 | 0.630 | 0.642 | 0.601 |
| Linear SVM | 0.689 | 0.666 | 0.689 | 0.674 |
| RBF kernel SVM | 0.632 | 0.545 | 0.632 | 0.573 |
| Logistic Regression | 0.660 | 0.560 | 0.660 | 0.598 |
| Multinomial Naive Bayes | 0.528 | 0.447 | 0.528 | 0.422 |
| MLP | 0.651 | 0.635 | 0.651 | 0.639 |
| CNN+BiLSTM | 0.708 | 0.692 | 0.708 | 0.695 |
| GPT-3.5 Simple Prompt | 0.400 | 0.442 | 0.391 | 0.414 |
| GPT-3.5 Complicated Prompt | 0.470 | 0.430 | 0.430 | 0.430 |
| **FSBM** | **0.717** | **0.702** | **0.716** | **0.708** |

Table 7: First stage results for balanced data.

| Classifiers/ Evaluation metrics | Accuracy | Precision | Recall | F-measure |
|---|---|---|---|---|
| Gaussian Naive Bayes | 0.765 | 0.749 | 0.765 | 0.754 |
| Decision Tree | 0.803 | 0.808 | 0.803 | 0.788 |
| **Gradient Boosting** | 0.850 | 0.848 | 0.850 | 0.844 |
| **Random Forest** | 0.864 | 0.859 | 0.864 | 0.859 |
| **Linear SVM** | 0.840 | 0.836 | 0.840 | 0.835 |
| **RBF kernel SVM** | 0.864 | 0.861 | 0.864 | 0.862 |
| **Logistic Regression** | 0.803 | 0.802 | 0.803 | 0.800 |
| **Multinomial Naive Bayes** | 0.690 | 0.716 | 0.690 | 0.683 |
| **MLP** | 0.850 | 0.846 | 0.850 | 0.845 |
| **CNN+ BiLSTM** | 0.873 | 0.873 | 0.870 | 0.869 |
| **GPT-3.5 Simple Prompt** | 0.434 | 0.401 | 0.510 | 0.448 |
| **GPT-3.5 Complicated Prompt** | 0.502 | 0.472 | 0.493 | 0.482 |
| **FSBM** | **0.891** | **0.881** | **0.885** | **0.882** |

The results for the second stage are presented in Tables 8 and 9, showcasing the outcomes for both imbalanced and balanced data, respectively. Consistently, the FSBM model outperformed other methods, demonstrating remarkable performance. Notably, the model achieved an accuracy of 94.1% for the balanced data scenario, representing the highest result among all conducted experiments.

Table 8: Second stage results for imbalanced data.

| **Classifiers/ Evaluation metrics** | **Accuracy** | **Precision** | **Recall** | **F-measure** |
|---|---|---|---|---|
| **Gaussian Naive Bayes** | 0.708 | 0.705 | 0.708 | 0.705 |
| **Decision Tree** | 0.717 | 0.720 | 0.717 | 0.718 |
| **Gradient Boosting** | 0.774 | 0.772 | 0.774 | 0.770 |
| **Random Forest** | 0.774 | 0.775 | 0.774 | 0.766 |
| **Linear SVM** | 0.745 | 0.742 | 0.745 | 0.742 |
| **RBF kernel SVM** | 0.755 | 0.757 | 0.755 | 0.745 |
| **Logistic Regression** | 0.774 | 0.782 | 0.774 | 0.762 |
| **Multinomial Naive Bayes** | 0.717 | 0.730 | 0.717 | 0.692 |
| **MLP** | 0.774 | 0.772 | 0.774 | 0.770 |
| **CNN+ BiLSTM** | 0.821 | 0.821 | 0.820 | 0.819 |
| **GPT-3.5 Simple Prompt** | 0.754 | 0.690 | 0.690 | 0.690 |
| **GPT-3.5 Complicated Prompt** | 0.773 | 0.666 | 0.857 | 0.749 |
| **FSBM** | **0.825** | **0.831** | **0.825** | **0.827** |

Table 9: Second stage results for balanced data.

| Classifiers Evaluation metrics | Accuracy | Precision | Recall | F-measure |
|---|---|---|---|---|
| **Gaussian Naive Bayes** | 0.883 | 0.895 | 0.883 | 0.882 |
| **Decision Tree** | 0.869 | 0.870 | 0.869 | 0.868 |
| **Gradient Boosting** | 0.887 | 0.890 | 0.887 | 0.887 |
| **Random Forest** | 0.901 | 0.906 | 0.901 | 0.901 |
| **Linear SVM** | 0.906 | 0.908 | 0.906 | 0.906 |
| **RBF kernel SVM** | 0.892 | 0.896 | 0.892 | 0.892 |
| **Logistic Regression** | 0.869 | 0.872 | 0.869 | 0.868 |
| **Multinomial Naive Bayes** | 0.869 | 0.873 | 0.869 | 0.868 |
| **MLP** | 0.901 | 0.902 | 0.901 | 0.901 |
| **CNN+ BiLSTM** | 0.932 | 0.933 | 0.932 | 0.932 |
| **GPT-3.5 Simple Prompt** | 0.780 | 0.760 | 0.730 | 0.744 |
| **GPT-3.5 Complicated Prompt** | 0.810 | 0.810 | 0.750 | 0.778 |
| **FSBM** | **0.941** | **0.941** | **0.942** | **0.941** |

Contrary to expectations, the LLM-based model (GPT-3.5) exhibited inferior performance, particularly when confronted with datasets comprising four distinct classes. This outcome can potentially be attributed to the general nature of the LLM approach, which may not adequately capture the specific and specialized nature of our medical dataset. In contrast, the applied method, which amalgamates deep learning and LLM concepts (ICL), capitalizes on the advantages of both approaches. By inheriting the shorter training time characteristic of LLMs and the high diagnostic accuracy associated with deep learning methods, the applied method achieved superior performance, yielding more precise and accurate results compared to the aforementioned techniques.

The results clearly indicate that the model performs significantly better when trained on balanced data compared to imbalanced data. Learning from imbalanced data poses inherent challenges [60], as the skewed distribution of training samples tends to bias standard classifiers in favor of the majority class, making it difficult to effectively recognize and classify rare instances [63]. In imbalanced data scenarios, there is a potential for rare minority samples to be erroneously treated as noise, while noise itself may be misidentified as minority samples [64]. Thus, the act of balancing the data has a positive impact on the performance of models and enhances the accuracy of prediction results.

Additionally, the second stage, which involves binary classification, exhibited superior results compared to the first stage, which focused on multiclass classification. Through the utilization of the second stage, the number of samples in each class is increased, thereby enabling algorithms to learn more effectively. Moreover, multiclass classification is commonly decomposed into binary classification, as it represents an efficient technique for decoding the classification process, known as class binarization [65]. This approach offers several advantages, including the relative ease of developing binary classifiers compared to multiclass classifiers [66]. Many classifiers inherently excel in binary classification tasks, displaying high performance [67]. In fact, numerous state-of-the-art classifiers are specifically designed for binary classification scenarios [68].

## V. CONCLUSION

Dental problems are prevalent worldwide, and delays or incorrect diagnoses can lead to inadequate treatment, negatively impacting oral health. Thus, this study focused on classifying the severity of dental diseases based on radiology reports. Results demonstrate that the used FSBM model outperforms other competing models in this context. Furthermore, balancing the dataset has a positive influence on the classification outcomes.

In future research, expanding the dataset size could lead to improved results. The effectiveness of such systems heavily relies on the availability of high-quality and well-annotated data, which remains a significant challenge in dentistry research. This study solely employed text data for training the classification model, but a multimodal approach incorporating both text and image data may yield more accurate results in future investigations. Additionally, the utilization of a GPT-4 model, capable of handling both text and image data, could be explored to enhance the classification process.

### Declarations

### Ethics approval and consent to participate

Not applicable.

### Consent for publication

Not applicable.

### Availability of data and materials

The dataset utilized in this study was obtained from a readily available source.

### Competing interests

The authors declare no competing interests.

### Funding

No Funding.